\documentclass{article}

\PassOptionsToPackage{numbers,sort&compress}{natbib}
\usepackage[dblblindworkshop,preprint]{neurips_2026}
\workshoptitle{Continual Learning for Enterprise AI Agents (CLEA)}

\usepackage[utf8]{inputenc}
\usepackage{hyperref}
\usepackage{url}
\usepackage{booktabs}
\usepackage{amsmath,amssymb,amsfonts}
\usepackage{xcolor}
\usepackage{graphicx}

\graphicspath{{figures/}}

\newcommand{\old}{\mathrm{old}}
\newcommand{\newc}{\mathrm{new}}

\newcommand{\stFT}{\textsc{Finetune}}
\newcommand{\stHEAD}{\textsc{FrozenHead}}
\newcommand{\stGLOBAL}{\textsc{Global}}
\newcommand{\stLOCAL}{\textsc{Local}}
\newcommand{\stDUAL}{\textsc{Dual}}

\newcommand{\blLwFNLam}{12}
\newcommand{\blLwFLamLo}{0.05}
\newcommand{\blLwFLamHi}{200}
\newcommand{\blEWCNLam}{6}
\newcommand{\blEWCLamLo}{$10^{8}$}
\newcommand{\blEWCLamHi}{$10^{14}$}
\newcommand{\nSeedsControlled}{5}

\newcommand{\nSeedsEtoE}{5}

\newcommand{\gapReplayOld}{0.022}

\newcommand{\gapHeadLo}{0.25}
\newcommand{\gapHeadHi}{0.28}
\newcommand{\resOldRecords}{36}
\newcommand{\resOld}{0.028}
\newcommand{\gapHeadRecords}{9}
\newcommand{\eeMinReplayOld}{0.868}
\newcommand{\eeGapReplayOld}{0.021}
\newcommand{\eeMaxReplayForget}{0.063}

\newcommand{\spreadReplayMax}{0.026}
\newcommand{\spreadOtherLo}{0.126}
\newcommand{\spreadOtherHi}{0.419}

\newcommand{\ctFTOld}{$0.560\!\pm\!0.027$}

\newcommand{\ctHEADOld}{$0.655\!\pm\!0.048$}

\newcommand{\ctGLOBALOld}{$0.909\!\pm\!0.048$}

\newcommand{\ctLOCALOld}{$0.924\!\pm\!0.042$}

\newcommand{\ctDUALOld}{$0.931\!\pm\!0.035$}
\newcommand{\ctDUALNew}{$0.905\!\pm\!0.041$}
\newcommand{\ctDUALForget}{$0.008\!\pm\!0.020$}
\newcommand{\blLwFOld}{$0.707\!\pm\!0.039$}
\newcommand{\blLwFNew}{$0.956\!\pm\!0.023$}

\newcommand{\blLwFLam}{0.25}
\newcommand{\blEWCOld}{$0.809\!\pm\!0.054$}
\newcommand{\blEWCNew}{$0.935\!\pm\!0.042$}

\newcommand{\blEWCLam}{$10^{11}$}

\newcommand{\eeFTOld}{$0.149\!\pm\!0.084$}

\newcommand{\eeLOCALOld}{$0.868\!\pm\!0.036$}

\newcommand{\eeDUALOld}{$0.889\!\pm\!0.055$}
\newcommand{\eeDUALNew}{$0.889\!\pm\!0.039$}

\newcommand{\nSeedsGrid}{5}
\newcommand{\gridNLo}{1}
\newcommand{\gridNHi}{6}
\newcommand{\gridBLo}{0}
\newcommand{\gridBHi}{6}
\newcommand{\gridNCells}{25}

\newcommand{\grnOnebThreeNew}{$0.556\!\pm\!0.104$}

\newcommand{\grnThreebThreeNew}{$0.800\!\pm\!0.115$}

\newcommand{\grnSixbZeroOld}{$0.253\!\pm\!0.083$}

\newcommand{\grnSixbOneOld}{$0.876\!\pm\!0.053$}

\newcommand{\grnSixbTwoOld}{$0.910\!\pm\!0.060$}

\newcommand{\grnSixbThreeNew}{$0.900\!\pm\!0.046$}

\newcommand{\gridNoRehearsalOld}{$0.513\!\pm\!0.209$}
\newcommand{\gridRefOld}{$0.889\!\pm\!0.059$}

\newcommand{\gridRefNew}{$0.978\!\pm\!0.030$}

\newcommand{\gridRefN}{24}
\newcommand{\gridRefB}{3}
\newcommand{\gridBestN}{6}
\newcommand{\gridBestB}{3}
\newcommand{\gridBestOld}{$0.917\!\pm\!0.065$}
\newcommand{\gridBestNew}{$0.900\!\pm\!0.046$}

\newcommand{\gridPlateauNLo}{3}
\newcommand{\gridPlateauBLo}{1}
\newcommand{\gridPlateauSpread}{0.054}
\newcommand{\nSeedsScreening}{5}
\newcommand{\detNKnown}{36}
\newcommand{\detNNovel}{72}
\newcommand{\scrQuantile}{0.9}
\newcommand{\detRecall}{$0.975\!\pm\!0.027$}
\newcommand{\detAccept}{$0.750\!\pm\!0.068$}

\newcommand{\detAuc}{$0.941\!\pm\!0.018$}

\newcommand{\scrMspAuc}{$0.884\!\pm\!0.033$}
\newcommand{\scrEnergyAuc}{$0.875\!\pm\!0.040$}
\newcommand{\scrOursAuc}{$0.941\!\pm\!0.018$}

\newcommand{\scrAccept}{0.89}
\newcommand{\scrGapMsp}{0.057}
\newcommand{\scrWinsMsp}{4}
\newcommand{\scrGapEnergy}{0.066}
\newcommand{\scrWinsEnergy}{5}

\title{When the Cross-Silo Federation Goes Offline:\\Continual Learning for Site Onboarding\\with Limited Unlabeled Data}

\author{%
  Ahmadreza Eslaminia \\
  Coordinated Science Laboratory \\
  University of Illinois Urbana-Champaign \\
  \texttt{ae15@illinois.edu} \\
  \And
  Klara Nahrstedt \\
  Coordinated Science Laboratory \\
  University of Illinois Urbana-Champaign \\
  \texttt{klara@illinois.edu} \\
  \And
  Chenhui Shao \\
  Dept.\ of Mechanical Engineering \\
  University of Michigan \\
  \texttt{chshao@umich.edu}
}

\begin{document}

\maketitle

\begin{abstract}
An organization often holds too little labeled data to train a model that
generalizes, and the records that would supply the rest sit with organizations
that cannot release them. Cross-silo federated learning offers a way through,
since participants exchange model parameters rather than records, but it
ordinarily settles two aspects of the arrangement in advance, the participating
sites and the classes the model can predict, and deployment can breach both. A new
site joins after training, once the established sites have finished their
engagement and gone offline, and its records arrive unlabeled, mixing conditions
the model already recognizes with conditions no participant has observed. We
present an autonomous three-stage procedure that expands the model entirely at the
joining site: reconstruction experts screen for novelty, clustering separates the
flagged records into candidate conditions, and class means describe the old
classes, all inside one shared representation. Those classes were learned from
records that never leave their owners, so the usual defenses against forgetting
are unavailable, and the procedure supplies the evidence they would have carried
from either of two dissimilar sources, prototypes held by the federation or
records held by the joining site. On a real industrial condition-monitoring
dataset, run end to end with no label consulted, either source holds old-class
accuracy at \eeMinReplayOld{} or above with forgetting at most
\eeMaxReplayForget{}, and the two differ by \eeGapReplayOld{}, less than a
single test record is worth, so a configuration can be chosen by the disclosure it
permits rather than the accuracy it delivers. Both keep old- and new-class accuracy
in balance where every alternative we measure gives up one task for the other, and
both retain more of the old classes than distillation- and regularization-based
baselines. The balance still holds with only \gridBestN{} labeled records per
arriving condition and \gridBestB{} retained per old class.

\end{abstract}

\section{Introduction}
\label{sec:introduction}

A single organization often holds too little labeled data to train a model that
generalizes well, while the records that would supply the rest sit with other
organizations that cannot release them, since regulation, contracts and
commercial sensitivity keep patient records, transaction histories and process
measurements inside the firm that collected them. Federated learning offers a way
through. The participants exchange model parameters instead of records, so each
contributes what its data taught without disclosing the data itself
\citep{mcmahan2017communication,kairouz2021advances}. Most deployments then settle
the arrangement at the outset in two respects. The set of participating sites is
fixed, and so is the set of classes the model can predict.

Federated continual learning relaxes these commitments
\citep{yoon2021federated,dong2022federated}, though usually under assumptions
that a site joining a model already in production is unlikely to meet.
Deployment can breach both at once. A new plant, hospital or business unit joins
after the model is in service and brings operating conditions the federation
never observed. By then the sites that trained the model may have completed their
engagement and gone offline, leaving the joining site what the base session
produced rather than a federation it can convene.
Class-incremental learning allows the output space to grow, but
training on the new conditions also degrades accuracy on the classes learned
earlier, an effect known as catastrophic forgetting \citep{masana2022class}.

\paragraph{Forgetting is harder to counter inside a federation.}
Centralized continual learning counters forgetting in three broad ways, and this
setting removes or weakens each. Replay stores a few examples of the old classes
and mixes them into the new training data \citep{rebuffi2017icarl}. It is usually
the most reliable and the first to fail here, because the old classes were learned
from records that belong to other organizations and never left them.
Regularization anchors the parameters to the previous solution using a curvature
estimate such as a Fisher information \citep{kirkpatrick2017ewc}; it survives, but
only if that estimate is computed where the old data are and shipped before the
sites disconnect. Distillation keeps the new model's outputs close to a frozen
copy of the old one \citep{li2018lwf}, and a joining site can apply it only to the
records it holds, so the old classes are anchored through data from a different
distribution.

\paragraph{What a late-joining site can and cannot use.}
Three constraints act together at the moment a site joins. Records stay where they
were collected, so the established sites contribute no exemplars and need not
remain reachable once the base model has been delivered. The arriving data carry
no labels, so the emerging conditions have to be discovered before the output
layer can grow. They are also scarce, since a recently joined site has little
operating history and annotating it consumes the expert effort the federation was
meant to conserve. What remains is one site, holding an unlabeled pool of tens of
examples per condition, that has to extend a model other organizations depend on.

\paragraph{One representation carries all three stages.}
The base federated session leaves behind more than a classifier. Its early layers
encode what the participating sites have in common, the structure of the
underlying process rather than the classes that session contained, and they are
identical at every site because the federation trained them together. We write
$\phi_\theta$ for the activation they produce and use it as the working space for
every later stage, scoring novelty there, clustering the flagged records there,
and describing old classes by their means in the same space. Since every stage
consumes activations rather than measurements, what leaves a site is a summary and
not a record.

\paragraph{Countering forgetting when the old data are elsewhere.}
Adding output units for the recovered conditions is mechanically simple. Keeping
the model accurate on the classes it already had, while it trains on the new ones,
is not. Something describing those classes has to enter the adaptation objective,
and two sources remain once the established sites have gone offline. The federation can leave behind class prototypes, or the joining site can
retain a small number of its own records, selected without labels by the same
screen that identifies novelty. The two differ far less in what they achieve than
in what they oblige an organization to disclose, so we treat them as a menu
indexed by the governing privacy constraint rather than as an ablation ladder.
Stated for a deployed system, the question is what a federation must carry
forward in order to keep serving its members, and which of the available carriers
a governance regime will permit.

\paragraph{Contributions.}
\begin{enumerate}\itemsep1pt
    \item An onboarding protocol for a site whose data are unlabeled and few, in
    which the incremental task is built autonomously at that site while the
    established sites stay offline, with screening, grouping and expansion all
    carried out inside one shared representation.
    \item An evaluation on real ultrasonic-welding process data covering
    different patterns of class arrival, against distillation- and
    regularization-based continual learning.
    \item A privacy-indexed account of old-class memory drawn from that
    evaluation. Federation-level prototypes and site-level records are dissimilar
    kinds of evidence, yet their accuracy differs by at most \gapReplayOld{}, so
    a deployment can choose between them by the disclosure it permits; both keep
    old-class and new-class accuracy within \spreadReplayMax{} of one another
    where every alternative we measure gives up one task to serve the other.
\end{enumerate}

\section{Related work}
\label{sec:related}

\paragraph{Federated and federated class-incremental learning.}
FedAvg established model averaging for decentralized training
\citep{mcmahan2017communication}, and later surveys catalogued the statistical,
systems and privacy problems that heterogeneous participants create
\citep{kairouz2021advances,li2020federated}. Federated class-incremental learning
admits classes that arrive over time. \citet{dong2022federated} showed that local
and global forgetting interact with client heterogeneity, a further line treats
the problem as one of selecting which participants' knowledge to accept rather
than which exemplars to store \citep{yoon2021federated}, and a recent survey
catalogues the task- and class-incremental variants
\citep{yang2026fclsurvey}. These protocols assume conditions the present setting
does not provide, namely labeled incremental tasks, clients that continue
participating in federated rounds, and a proxy server, per-client exemplar memory
or server-side generator \citep{dong2022federated,zhang2023target}. Here a single
organization arrives with an unlabeled pool and the established sites do not
reconnect, so we compare against the centralized continual-learning methods that
the same literature adapts when a federated protocol does not transfer
\citep{zhang2023target}.

\paragraph{Open-set recognition and replay.}
Open-set recognition requires a classifier to reject inputs outside its label set
\citep{scheirer2013toward,bendale2016towards}, and reconstruction models suit the
case in which examples of future classes cannot exist at training time
\citep{chandola2009anomaly}. The federated version has been addressed through
disagreement between clients \citep{zhu2024fedoss}, which assumes the clients are
online at the same time. We train one federated expert per known class, and the
same screen also selects the records that anchor the old classes. Replay is among
the most dependable defenses against forgetting
\citep{rebuffi2017icarl,buzzega2020dark,masana2022class}, and its exemplars are
normally chosen using labels, whereas prototypes are summaries obtained by
averaging.

\paragraph{Discovering the classes before learning them.}
A smaller body of work asks a federation to name classes that no site has labeled.
Federated generalized category discovery clusters unlabeled client data into known
and novel categories, but treats discovery as the end of the task rather than the
start of an expansion \citep{pu2024fedgcd}, and federated open-world
semi-supervised learning spans both halves with clients holding unlabeled data
that contains novel classes \citep{zhang2023fedopenworld}. Federated continual
novel class learning is the closest published setting to ours that we are aware
of: its clients detect novel classes, the server aligns their identity across
sites, and training continues so that earlier classes survive
\citep{wang2024fedcn}. Two conditions separate our problem from it: discovery
there proceeds while every site still participates, whereas here the established
sites are permanently offline and only a model and a small set of class summaries
outlive them; and its clients hold enough unlabeled data for cross-client
agreement to identify a novel class, while the site arriving here holds a few
dozen records in total and must decide alone. Few-shot class-incremental learning
addresses the second condition in the centralized case, typically by freezing the
representation after a base session and placing new classes with prototypes
\citep{wang2023teen}, but receives those classes already named and labeled.

\paragraph{Cross-silo federations.}
The setting studied here is cross-silo rather than cross-device, meaning a small
number of organizations, each holding data it is legally or contractually unable
to release, and each participating under a persistent identity
\citep{kairouz2021advances,li2020federated}. Hospitals, banks and manufacturers
all train shared models over records they cannot release to one another
\citep{rieke2020future,long2020openbanking,berghout2023federated,becker2022federated,eslaminia2024fdgcm}.
Two features
recur and shape the present problem: labeled data are scarce at any single silo,
and the sensitivity of the records is the reason the federation exists at all, so
every quantity a site releases has to be justified. We instantiate the protocol on
ultrasonic welding, where unknown-fault detection and few-shot adaptation have
been studied for a single machine and a single dataset
\citep{eslaminia2026adaptive}. That setting is centralized, so it faces neither
the disclosure constraint nor the offline-participant constraint that shape the
problem here.

\section{Problem formulation}
\label{sec:setting}

\subsection{Setting}

Sites $1,\ldots,M$ hold private datasets over a class set $\mathcal{Y}_{\old}$ of
size $K$, and a base federated session learns $f_{\theta}^{0}$ without moving raw
records. After deployment, site $M{+}1$ joins with an unlabeled pool
$\mathcal{U}$ containing a mixture of $\mathcal{Y}_{\old}$ and a disjoint set
$\mathcal{Y}_{\newc}$ whose size and identity are unknown on arrival. The
classifier expands to $K+R$ outputs using only what is available at the joining
site together with the summaries the federation retained.

The federation retains the shared model, $K$ reconstruction experts, the
detection constants defined in Section~\ref{sec:method}, and one
representation-level prototype per site and old class. The joining site keeps up
to $b$ of its own records per old class when the deployment permits it.
Adaptation runs at site $M{+}1$ alone, so the established sites need not
reconnect and none of their records are transferred.

\subsection{Evaluation protocol}

We evaluate on one common held-out test set and report three quantities. Before
the new site joins, $A_{\old}^{0}$ is the old-class accuracy of the base model,
averaged over the $M$ established sites. After adaptation, $A_{\old}$ is the
old-class accuracy of the expanded model averaged over those same $M$ sites, and
$A_{\newc}$ is its accuracy on the conditions that arrived. There is one expanded
model and one test set, and the per-site figures differ only in that each site
reads those records through its own local input statistics. Forgetting is the drop
on that same population, $F_{\old}=A_{\old}^{0}-A_{\old}$, so it measures what
the established sites lose and the joining site never enters it. That site's own
old-class accuracy is reported separately, since no before-adaptation figure
exists for it, and reporting the old classes before and after separates what the
federation loses from what it gains on the new task.

\section{Method}
\label{sec:method}

\begin{figure}[t]
    \centering
    \includegraphics[width=0.85\linewidth]{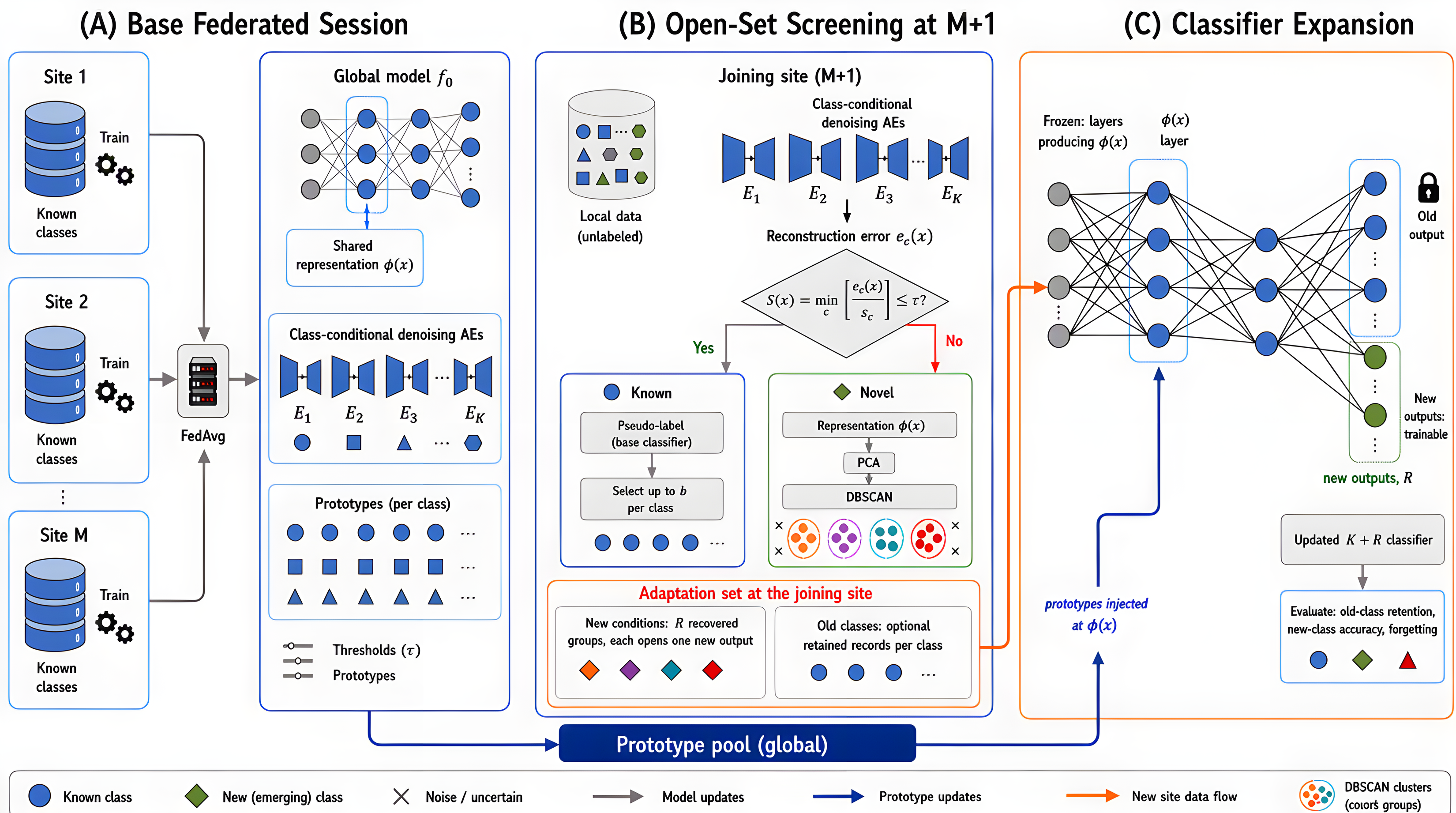}
    \caption{Onboarding a late-joining site. (A) The base federated session leaves
    behind a shared classifier, one reconstruction expert per known class, and
    representation-level prototypes. (B) The joining site screens its own
    unlabeled pool against those experts and clusters what it flags into
    candidate conditions. (C) The classifier expands by one output per recovered
    group, with the layers producing $\phi_\theta$ held fixed. Prototype replay
    and local rehearsal are alternative sources of old-class evidence.}
    \label{fig:pipeline}
\end{figure}

The protocol performs three stages at the joining site. It screens the unlabeled pool
for novelty, groups the flagged records into candidate conditions, and expands the
classifier under an old-class memory constraint. Every stage operates on the same
hidden representation, which is what allows the procedure to run without labels.
Figure~\ref{fig:pipeline} identifies the quantities that cross a site boundary.

\subsection{The shared representation}

Each site trains a multilayer perceptron locally and the server averages the local
parameters weighted by sample count. Let $\phi_\theta(x)$ be the activation of an
early hidden layer of the resulting model. How many layers precede $\phi_\theta$
is a hyperparameter of the protocol: deep enough that $\phi_\theta$ no longer
carries the input variables themselves, and shallow enough that what it encodes
belongs to the process the sites share rather than to the classes the base session
contained, which is what allows it to represent conditions absent at training
time.

Each stage relies on a different property of this space. Screening needs one in which a
record's dissimilarity to every known class can be measured, and the base session
already shapes $\phi_\theta$ so that the known classes separate in it. Grouping
needs a space of manageable dimension in which records of one condition lie close
together, and the same representation supplies one without a second embedding that
the joining site has no labels to validate. Replay needs a form in which an
established site can describe an old class without transmitting one of its
records, and a mean activation serves, since every site reads this space through
the same parameters while the mean itself is a summary rather than anything the
site measured.

\subsection{Screening: identifying unfamiliar records}

For each $c\in\mathcal{Y}_{\old}$ the established sites jointly train a denoising
autoencoder $g_c$ on representations labeled $c$ only, using Gaussian input noise
and the clean representation as the target. The server averages each expert
independently. At inference,
\begin{equation}
e_c(x)=\left\|\phi_\theta(x)-g_c\!\left(\phi_\theta(x)\right)\right\|_2^2,
\qquad
S(x)=\min_{c\in\mathcal{Y}_{\old}} \frac{e_c(x)}{s_c}.
\end{equation}
The score has a simple interpretation. A familiar record is reconstructed well by
one specialist, whereas an unfamiliar record is explained by none of them. Using
one expert per class rather than a single global model is what makes this
interpretation available, since an autoencoder fitted to all known classes
averages over a multimodal distribution.

The divisor $s_c$ puts the experts on a common scale. Classes are not equally easy
to reconstruct, so an unnormalized minimum is governed by whichever expert fits
most loosely. Site $m$ reports the median error of expert $c$ on its own class-$c$
records, then a high quantile of its own known-class scores, and the server
averages each across sites,
\begin{equation}
s_c=\frac{1}{M}\sum_{m=1}^{M}
\operatorname{med}\left\{e_c(x):x\in\mathcal{D}_m,\; y=c\right\},
\qquad
\tau=\frac{1}{M}\sum_{m=1}^{M}Q_q\!\left(\{S(x):x\in\mathcal{D}_m\}\right).
\end{equation}
Each expert's error is thereby judged against what it normally achieves on the
class it owns, and the joining site flags $x$ when $S(x)>\tau$. Neither constant
uses an example of a new class: each site releases $K$ medians of its own
known-class errors and one quantile of its own known-class scores, $K{+}1$ scalars
beyond the model updates.

\subsection{Grouping: estimating how many conditions have arrived}

Screening identifies unfamiliar records but not how many conditions they
represent. Their frozen representations are reduced by principal component
analysis and clustered with DBSCAN \citep{ester1996density}, which needs no target
cluster count and leaves low-density points unassigned. Each recovered group opens
one new output, and no label is consulted.

\subsection{Expansion and the evidence it requires}

Once $R$ groups exist, the output layer grows from $K$ to $K+R$ units. The weight
vector and bias that produce each of the $K$ old-class logits are copied from the
base model and their gradients are set to zero, so the linear map from the
representation to the old outputs survives adaptation unchanged. Every
configuration below keeps this masking except \stFT{}, which is plain
fine-tuning.

Preserving that map is not enough to keep the old classes, because nothing in the
adaptation objective refers to them. Containing forgetting needs
evidence about the old classes inside that objective, and two sources remain
available. They differ in what each obliges an organization to reveal.

\paragraph{Federation-level evidence: prototypes.}
Before the established sites disconnect, the base federation stores one prototype
per site and old class,
\begin{equation}
p_{m,c}=\frac{1}{|\mathcal{D}_{m,c}|}\sum_{(x,y)\in\mathcal{D}_{m,c}}\phi_{\theta^0}(x),
\end{equation}
the mean activation of that site's records for that class. The average is taken in
the representation space deliberately. A mean of raw measurements would carry
recognizable process values, whereas a mean activation lives in a space defined by
parameters the federation already holds, so a site describes an old class without
releasing anything it measured, and these $MK$ vectors outlive the sites that
produced them. During adaptation each
prototype is injected at the frozen representation layer, forwarded through the
remaining classifier, and its cross-entropy added to the loss on real data,
\begin{equation}
\mathcal{L}_{\mathrm{inc}}
=\mathcal{L}_{\mathrm{CE}}(x,y)
+\lambda_{\mathrm{p}}\,\mathcal{L}_{\mathrm{CE}}^{\mathrm{proto}}(p_{m,c},c).
\end{equation}

\paragraph{Site-level evidence: local rehearsal.}
The joining site can instead assemble its own anchor set without labels, because
the screen that identifies novelty also selects it. Records with $S(x)\leq\tau$ are
pseudo-labeled by the frozen base classifier and the site retains up to $b$ per
predicted old class, drawn uniformly. These records are never transmitted and the
established sites supply no class means.

\subsection{A menu indexed by disclosure}

Table~\ref{tab:strategies} lists the configurations. They are not successive steps
toward one best setting; each answers differently what a deployment may release,
and Section~\ref{sec:results} reports the accuracy each obtains.

\begin{table}[t]
\caption{Adaptation configurations, what each holds fixed, and the disclosure each
requires. $b$ is the number of records the joining site retains per old class.
Freezing $\phi_\theta$ keeps prototypes and retained records in a coordinate
system that does not shift while the classifier is expanded.}
\label{tab:strategies}
\centering
\footnotesize
\begin{tabular}{lccccl}
\toprule
& \multicolumn{2}{c}{Held fixed} & \multicolumn{2}{c}{Old-class evidence} & \\
\cmidrule(lr){2-3}\cmidrule(lr){4-5}
Name & Old outputs & $\phi_\theta$ & Prototypes & Records & Applicable when \\
\midrule
\stFT     & no  & no  & no  & no        & no old-class evidence is available \\
\stHEAD   & yes & no  & no  & no        & only the model may be reused \\
\stGLOBAL & yes & yes & yes & no        & class means may be released, nothing retained \\
\stLOCAL  & yes & yes & no  & $b$/class & nothing released, records retained locally \\
\stDUAL   & yes & yes & yes & $b$/class & both disclosures are permitted \\
\bottomrule
\end{tabular}
\end{table}

\section{Experimental setup}
\label{sec:experiments}

\paragraph{Instantiation.}
The protocol is instantiated on ultrasonic metal welding, a joining process used
in battery and electronics manufacturing. The data comprise 270 observations, one per
welding cycle, each a vector of 252 engineered signal features recorded on
production equipment, organized into nine balanced classes formed by three tool states (new, worn, damaged) crossed
with three surface states. The new- and worn-tool combinations are the six old
classes, and the three damaged-tool combinations arrive later. Tool damage is a
realistic condition to arrive late, since it develops with use and need not have
occurred anywhere in the federation while the base model was trained. It is also
the demanding choice, since the arriving classes pair damaged tools with surface
states the federation already knows and so differ from familiar conditions along
one factor only.

The 30 records of each class are divided so that no observation is used twice.
Six form a common test set on which no stage trains, six go to each of the three
established sites for the old classes, and the remainder go to the joining site:
six per old class and 24 per new class. The joining site therefore receives a
pool of 36 familiar and 72 unfamiliar records, and every accuracy in the paper is
measured on the 54-record test set. The data scale reflects the operating regime,
since process data of this kind are proprietary, costly to label, and accumulate
slowly at any one site.

\paragraph{Models and optimization.}
The base classifier is a multilayer perceptron with hidden widths 150, 90 and 40,
trained for 20 federated rounds. Here $\phi_\theta$ is the activation of the
first of those layers and is therefore 150-dimensional.
One denoising autoencoder per old class operates in that space, and the detection
quantile is $q=0.95$. Adaptation runs 20 site-local rounds with prototype loss
weight one and $b=3$ retained records per old class, so federation-level memory
comprises $3\times6=18$ prototypes of dimension 150.
Appendix~\ref{app:reproducibility} gives the remaining hyperparameters.

\paragraph{Reference methods.}
Two established continual-learning methods are evaluated under the same
architecture, split, optimization schedule and class-arrival patterns.
\textsc{LwF} distils the frozen base model's old-class outputs on the joining
site's adaptation records \citep{li2018lwf}, and \textsc{EWC} penalizes parameter
movement weighted by a diagonal Fisher information computed at the established
sites and averaged by the server \citep{kirkpatrick2017ewc}. Both are centralized
methods that the federated class-incremental literature adapts as references when
its own protocols do not transfer \citep{zhang2023target}. Neither has a default
weight that carries over here, so each is reported at the most favorable balanced
setting it reaches in a sweep, which favors the reference methods; our own
configurations use a fixed prototype weight of one and are not swept
(Appendix~\ref{app:sweep}).

\paragraph{What we measure.}
We run two main experiments. The first isolates the expansion step. New classes are
given with verified labels, so screening and grouping contribute no error and the
only variable is which old-class evidence a configuration replays. Every
configuration is evaluated on all seven non-empty subsets of the three new
classes, exercising arrivals of one, two and three classes, over
\nSeedsControlled{} seeded splits, and Table~\ref{tab:controlled}(a) reports it.
The second runs the procedure as deployed. The joining site's pool is screened,
grouped and used for expansion with no label consulted, so early errors propagate
forward, and Table~\ref{tab:controlled}(b) reports it. Screening is also reported
separately on the same pool.

A jointly trained nine-class model is not an available reference, since it would
need simultaneous access to historical and joining-site records, so base-session
accuracy serves as the reference instead. The two panels run separate base
sessions, whose per-seed accuracies agree to within one test record. Reported
spreads are sample standard deviations across seeds. Old-class accuracy is measured on the
\resOldRecords{} old-class records of the test set, so one record is worth
\resOld{}, and we do not read a difference below that as separating two
configurations.

\section{Results}
\label{sec:results}

\subsection{Screening}

On the pool the joining site receives, \detNKnown{} familiar and \detNNovel{}
unfamiliar records, the detector reaches recall \detRecall{} and acceptance
\detAccept{} at the federated threshold, with ROC-AUC \detAuc{}; on average it
misses under two unfamiliar records and flags one familiar record in four in
error. High recall is the right operating point for
onboarding, since a condition that is never flagged is never learned, whereas a
familiar record flagged in error is passed to grouping, where it is either left
unassigned or absorbed into a recovered group.
Appendix~\ref{app:screening} compares this score against post-hoc detectors read
off the same classifier.

\subsection{The evidence that expansion requires}

Table~\ref{tab:controlled}(a) attributes old-class accuracy to the evidence that
produces it, averaging each seed over all seven patterns of class arrival and using
verified new-class labels so that screening errors are excluded. Two
configurations lose the old classes, fine-tuning falling to \ctFTOld{} and
\stHEAD{} reaching \ctHEADOld{}. Neither carries any description of the
old classes into adaptation, and both forget accordingly.

Both forms of replay restore old-class accuracy to comparable levels.
\stGLOBAL{} reaches \ctGLOBALOld{}, \stLOCAL{} reaches \ctLOCALOld{}, and
\stDUAL{} reaches \ctDUALOld{} with forgetting of \ctDUALForget. The three sit
inside a band this experiment cannot resolve. The largest separation among them
is \gapReplayOld{}, smaller than the \resOld{} a single test record is worth,
whereas each is separated from \stHEAD{} alone by \gapHeadLo{} to
\gapHeadHi{}, which is \gapHeadRecords{} test records or more. We therefore treat the
three as equivalent in accuracy and separate them only by what they oblige a
participant to disclose.

\begin{table}[t]
\caption{Accuracy by configuration, over \nSeedsControlled{} splits. Old-class
accuracy is given separately for the established sites and the joining site; the
base session precedes the join and has no joining-site entry. Forgetting is the
drop in established-site old-class accuracy, computed on those sites alone.
(a) Controlled arrivals, averaged within each split over all seven arrival
patterns; the lower block gives the baselines at their best regularization
weight. (b) The full autonomous procedure.}
\label{tab:controlled}
\centering
\footnotesize
(a) Controlled arrivals, verified new-class labels\\[2pt]
\begin{tabular}{lcccc}
\toprule
Configuration & Old (est.\ sites) & Old (joining) & New classes & Forgetting \\
\midrule
Base session & $0.939\!\pm\!0.023$ & & & \\
\midrule
\stFT & $0.560\!\pm\!0.027$ & $0.567\!\pm\!0.027$ & $0.979\!\pm\!0.010$ & $0.379\!\pm\!0.037$ \\
\stHEAD & $0.655\!\pm\!0.048$ & $0.658\!\pm\!0.045$ & $0.979\!\pm\!0.013$ & $0.284\!\pm\!0.060$ \\
\stGLOBAL & $0.909\!\pm\!0.048$ & $0.913\!\pm\!0.048$ & $0.909\!\pm\!0.060$ & $0.030\!\pm\!0.026$ \\
\stLOCAL & $0.924\!\pm\!0.042$ & $0.927\!\pm\!0.044$ & $0.906\!\pm\!0.039$ & $0.015\!\pm\!0.029$ \\
\stDUAL & $0.931\!\pm\!0.035$ & $0.933\!\pm\!0.035$ & $0.905\!\pm\!0.041$ & $0.008\!\pm\!0.020$ \\
\midrule
\textsc{LwF} & $0.707\!\pm\!0.039$ & $0.716\!\pm\!0.037$ & $0.956\!\pm\!0.023$ & $0.232\!\pm\!0.042$ \\
\textsc{EWC} & $0.809\!\pm\!0.054$ & $0.812\!\pm\!0.054$ & $0.935\!\pm\!0.042$ & $0.130\!\pm\!0.041$ \\
\bottomrule
\end{tabular}

\\[5pt]
(b) Full autonomous procedure\\[2pt]
\begin{tabular}{lcccc}
\toprule
Configuration & Old (est.\ sites) & Old (joining) & New classes & Forgetting \\
\midrule
Base session & $0.931\!\pm\!0.025$ & & & \\
\midrule
\stFT & $0.149\!\pm\!0.084$ & $0.144\!\pm\!0.087$ & $0.944\!\pm\!0.039$ & $0.783\!\pm\!0.066$ \\
\stHEAD & $0.311\!\pm\!0.170$ & $0.311\!\pm\!0.161$ & $0.956\!\pm\!0.025$ & $0.620\!\pm\!0.170$ \\
\stGLOBAL & $0.872\!\pm\!0.056$ & $0.872\!\pm\!0.064$ & $0.878\!\pm\!0.099$ & $0.059\!\pm\!0.059$ \\
\stLOCAL & $0.868\!\pm\!0.036$ & $0.872\!\pm\!0.037$ & $0.900\!\pm\!0.046$ & $0.063\!\pm\!0.037$ \\
\stDUAL & $0.889\!\pm\!0.055$ & $0.889\!\pm\!0.068$ & $0.889\!\pm\!0.039$ & $0.043\!\pm\!0.042$ \\
\bottomrule
\end{tabular}

\end{table}

The lower block of panel (a) places these results beside established
continual-learning practice. Both reference methods restrain movement away from
the previous solution and control that restraint with a single scalar, so sweeping
it moves accuracy along a frontier rather than raising it. At their balanced
operating points \textsc{LwF} reaches \blLwFOld{} old-class accuracy with
\blLwFNew{} on the new conditions, and \textsc{EWC} reaches \blEWCOld{} with
\blEWCNew{} (Appendix~\ref{app:sweep}). The replay
configurations lie on neither frontier, with \stDUAL{} sustaining \ctDUALOld{}
and \ctDUALNew{} at once. Both reference methods obtain higher new-class accuracy
and markedly lower old-class accuracy, so the difference lies in retention rather
than plasticity, and supplying evidence about the old classes is not equivalent to
restraining movement away from them.

Replay also balances the two tasks more evenly. All three replay configurations
keep old- and new-class accuracy within \spreadReplayMax{} of one another, whereas
every other entry in the table separates them by \spreadOtherLo{} to
\spreadOtherHi. Appendix~\ref{app:budget} varies the joining site's data budget
under \stLOCAL{} and finds the same balance at \gridBestN{} labeled records per
arriving condition and \gridBestB{} retained per old class.

\subsection{The procedure end to end}

Panel (b) of Table~\ref{tab:controlled} chains every stage. The pool is screened,
the flagged records are grouped, each group opens an output, and adaptation proceeds
with no label consulted, so errors propagate into the new-class training set and,
through pseudo-labeling, the old-class rehearsal buffer. A recovered group carries
no name, so each is matched to the condition holding most of its members; in every
split that match was one-to-one. Appendix~\ref{app:seeds} lists each split's
grouping outcome.

After autonomous expansion, \stDUAL{} retains \eeDUALOld{} on the old classes
while reaching \eeDUALNew{} on conditions it discovered without supervision,
against \eeFTOld{} for fine-tuning. Retention is within a few points of the
controlled setting despite the propagated errors, and the three replay
configurations again fall inside one band, from \eeLOCALOld{} to \eeDUALOld{}.
A rehearsal buffer assembled from predictions admits screening errors that
prototypes computed before deployment do not, which is a reason to expect the two
to diverge here, and these runs do not show it.

\section{Selecting a configuration}
\label{sec:deployment}

The margins in Section~\ref{sec:results} do not support reading
Table~\ref{tab:strategies} as an ablation ladder. What separates the replay
configurations in practice is not accuracy but permission, and the two options ask
permission of different parties.
\stGLOBAL{} needs the established sites to release $MK$ class means and needs
nothing of the joining site, whereas \stLOCAL{} asks the established sites
for no class means, needing only the model and the screening constants they
publish once, and asks the joining site to retain a few of its own records. Neither requirement implies the other, and a deployment ordinarily finds
one foreclosed. Where an organization may publish representation-level statistics
but the joining site is bound by a retention policy, \stGLOBAL{} applies. Where
the established sites will release no class means, the common position
between separate legal entities, \stLOCAL{} reaches comparable accuracy from
records that never leave the joining site, and \stDUAL{} applies where both
disclosures are permitted. A configuration can therefore be chosen from the
binding constraint, with its cost known in advance.

\section{Limitations and future work}
\label{sec:limitations}

\paragraph{A fixed and common feature space.}
The protocol assumes that every site measures the same quantities in the same way
and that this does not change while the federation runs. A prototype is
interpretable at another site only because both compute $\phi_\theta$ from the
same inputs with the same parameters, and the threshold is comparable across
sites for the same reason. Re-instrumenting a line, replacing a sensor or
altering a feature-extraction pipeline breaks that correspondence, and a site with
a partly different feature set cannot be admitted as the procedure stands.
Aligning heterogeneous feature spaces beforehand is the extension we would attempt
first.

\paragraph{Balanced classes and similar sites.}
Every class here contributes the same number of records and every site observes
one process on one machine, so two pressures a real federation applies are absent.
Rare conditions would arrive as a handful of records, moving the operating point
of the screen, and sites differing by machine or operator would shift the
representation the prototypes summarize.

\paragraph{Scope of the study.}
The federation is assumed to share one classification task with a fixed label
space, and the protocol expands that space once per joining site. Sites carrying
related but distinct tasks, and longer sequences in which the process drifts
between arrivals, lie outside what we measure. Screening and grouping act on a
representation and would carry over to other task families, whereas expansion is
stated in terms of output units and would have to be restated for a regression or
segmentation head, and the reported effect sizes are properties of one industrial
instantiation. The reference methods are also compared as published, so neither
receives the masking, freezing or retained records the replay configurations use.
A fuller study would evaluate each stage against its own alternatives, screening,
grouping and expansion in turn, and then assemble the strongest option at each
stage into the autonomous procedure; that is the comparison we would run next.

\section{Conclusion}
\label{sec:conclusion}

We examined what a deployed federated model needs to absorb a site whose data are
unlabeled and contain conditions no participant has observed. The protocol does
this inside one shared representation, at the joining site, with
no label consulted and no established site online. Containing forgetting requires evidence
about the old classes inside the adaptation objective, and two dissimilar sources
supply it at comparable accuracy, so a deployment can choose by disclosure rather
than performance.

\begin{ack}
We thank Prof.\ Placid Ferreira for his advice and guidance throughout this work.
\end{ack}

\newpage
\bibliographystyle{plainnat}
{\small\bibliography{references}}

\newpage
\appendix
\renewcommand{\thetable}{A\arabic{table}}
\renewcommand{\thefigure}{A\arabic{figure}}
\renewcommand{\theHtable}{A\arabic{table}}
\renewcommand{\theHfigure}{A\arabic{figure}}
\setcounter{table}{0}
\setcounter{figure}{0}
\section{Screening}
\label{app:screening}

Screening is a supporting stage of the protocol, and the memory results do not
depend on which detector supplies the candidate set. Table~\ref{tab:screening}
compares the class-conditional reconstruction score against the two post-hoc
scores that the joining site can compute from the shared model alone, the maximum
softmax probability \citep{hendrycks2017baseline} and the energy score
\citep{liu2020energy}, neither of which needs a further statistic released by the
established sites. All three are evaluated on
the same \detNKnown{} familiar and \detNNovel{} unfamiliar records over
\nSeedsScreening{} seeds.

Both classifier-confidence scores are weaker here, reaching \scrMspAuc{} and
\scrEnergyAuc{} ROC-AUC against \scrOursAuc{} for the reconstruction bank, which
is consistent with a six-class softmax trained on 36 observations per site being
poorly calibrated. The margins, \scrGapMsp{} and \scrGapEnergy{}, are larger
than any separation among the memory configurations of
Table~\ref{tab:controlled}, and the bank ranks ahead of the energy score on
\scrWinsEnergy{} of \nSeedsScreening{} splits and ahead of maximum softmax
probability on \scrWinsMsp{} of \nSeedsScreening{}. What the bank adds beyond
the margin is a per-class account of why a record was flagged, since the errors
of the individual experts are available alongside the decision.

\begin{table}[ht]
\caption{Post-hoc screening scores on the joining site's pool, over
\nSeedsScreening{} seeds. Recall is measured with each score thresholded at the
\scrQuantile{} quantile of its own familiar-record scores, so all three operate at
the same acceptance rate, \scrAccept. Mean $\pm$ sample standard deviation.}
\label{tab:screening}
\centering
\small
\begin{tabular}{lccc}
\toprule
Score & ROC-AUC & Average precision & Recall \\
\midrule
Maximum softmax probability & $0.884\!\pm\!0.033$ & $0.923\!\pm\!0.032$ & $0.614\!\pm\!0.085$ \\
Energy score & $0.875\!\pm\!0.040$ & $0.932\!\pm\!0.020$ & $0.636\!\pm\!0.095$ \\
Class-conditional reconstruction (ours) & $0.941\!\pm\!0.018$ & $0.950\!\pm\!0.034$ & $0.872\!\pm\!0.057$ \\
\bottomrule
\end{tabular}

\end{table}

\section{Reference-method sweeps}
\label{app:sweep}

\begin{figure}[ht]
    \centering
    \includegraphics[width=0.68\linewidth]{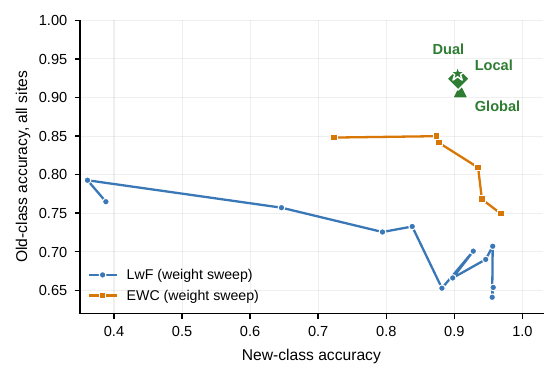}
    \caption{Regularization sweeps for the reference methods against the
    replay-based configurations, \nSeedsControlled{} seeds. Each curve traces one
    method as its regularization weight increases. Old-class accuracy is bought
    with new-class accuracy along each sweep; the replay configurations lie on
    neither.}
    \label{fig:sweep}
\end{figure}

Figure~\ref{fig:sweep} reports the full regularization sweep for both reference
methods, \blLwFNLam{} weights for \textsc{LwF} spanning \blLwFLamLo{} to
\blLwFLamHi{} and \blEWCNLam{} for \textsc{EWC} spanning \blEWCLamLo{} to
\blEWCLamHi, each averaged over \nSeedsControlled{} seeds. Both were extended
until the harmonic mean of the two accuracies peaked and began to decline, so
neither method is reported at an under-tuned setting; the rows in
Table~\ref{tab:controlled} correspond to weights \blLwFLam{} and \blEWCLam{}.
That balanced point favors the reference methods, which are the only entries in
the table selected this way; our own configurations use a fixed prototype weight
and are not swept.

The sweep is presented rather than a single row because any choice of operating
point is arguable, and the figure makes the comparison independent of that choice.
The replay configurations sit above both sweeps at every weight. At comparable
new-class accuracy they retain substantially more, and no setting of either weight
reaches the region they occupy.

\section{How much data the joining site needs}
\label{app:budget}

Both kinds of evidence that adaptation consumes come from the joining site,
namely the labeled records that define the arriving conditions and the records it
retains for the old ones. Neither is plentiful at a site that has recently
started operating, so the question a deployment asks is not how the protocol
behaves at the budget used in Section~\ref{sec:experiments} but how far that
budget can be cut. We answer it in the \stLOCAL{} configuration, where no
prototype is available and every training record therefore comes from the joining
site, over a grid of \gridNCells{} budgets: $n$ from \gridNLo{} to \gridNHi{}
labeled records per arriving condition, and $b$ from \gridBLo{} to \gridBHi{}
retained records per old class. All three conditions arrive at once, every other
setting is that of Section~\ref{sec:experiments}, and each cell is averaged over
\nSeedsGrid{} splits.

\begin{figure}[ht]
    \centering
    \includegraphics[width=\linewidth]{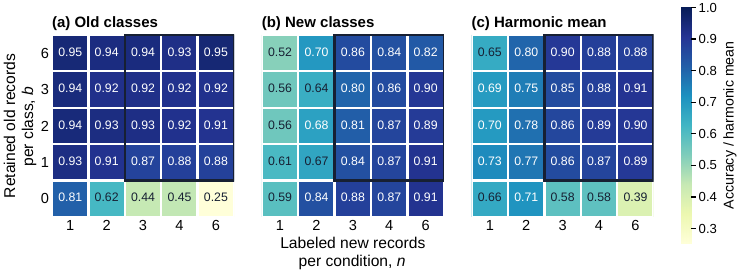}
    \caption{Joining-site data budget under \stLOCAL, over \nSeedsGrid{} splits
    with all three conditions arriving at once. $n$ is the number of labeled
    records per arriving condition and $b$ the number retained per old class.
    (a) old-class accuracy over all participating sites and (b) new-class accuracy
    at the joining site are the two quantities Table~\ref{tab:controlled} reports;
    (c) is their harmonic mean, which is high only where both are. The outlined
    block marks the region $n\geq\gridPlateauNLo{}$, $b\geq\gridPlateauBLo{}$
    discussed in the text.}
    \label{fig:budget}
\end{figure}

Figure~\ref{fig:budget} reports the grid, and the two budgets separate along the
two axes.

\paragraph{Arriving-condition records buy plasticity.}
New-class accuracy, panel (b), is governed by $n$ and is nearly flat in $b$. One
labeled record per condition leaves it at \grnOnebThreeNew{} at the default
$b=3$, since a single point fixes an output unit without describing the region
around it. Three records reach \grnThreebThreeNew{} and six reach
\grnSixbThreeNew{}, so most of the plasticity the site needs is bought within the
first handful of examples.

\paragraph{Retained records buy retention, and the first one does most of it.}
Old-class accuracy, panel (a), is governed by $b$. The $b=\gridBLo{}$ row is
adaptation with no old-class evidence at all, and it is the only region of the
grid where the model loses the old classes, averaging \gridNoRehearsalOld{}
across that row. It also degrades as $n$ grows, falling to \grnSixbZeroOld{} at
$n=\gridNHi{}$, which is the mechanism of Section~\ref{sec:results} seen
directly: the more the site trains on the arriving conditions, the further the
shared layers move and the less the preserved output map means. A single retained
record per class lifts that column to \grnSixbOneOld{}, a second reaches
\grnSixbTwoOld{}, and the rows above change little.

\paragraph{The cheapest adequate budget.}
Panel (c) is high only where both accuracies are, and it shows a broad plateau
rather than a single optimum. Every cell with $n\geq\gridPlateauNLo{}$ and
$b\geq\gridPlateauBLo{}$ lies within \gridPlateauSpread{} of the best cell in the
grid, $n=\gridBestN{}$ with $b=\gridBestB{}$, at \gridBestOld{} on the old
classes and \gridBestNew{} on the new ones. Outside that region one of the two
gives way, the old classes at $b=\gridBLo{}$ and the new ones below
$n=\gridPlateauNLo{}$. The budget the main experiments use is far larger,
$n=\gridRefN{}$ and $b=\gridRefB{}$, and reaches \gridRefOld{} and \gridRefNew{}
under the same configuration and splits, so the largest cell in the grid is
already above the full-pool budget on the old classes and below it on the new
ones. Reducing the arriving-condition budget costs accuracy on the conditions the
site is itself trying to learn and returns a little on the classes the federation
already served, which is the more forgiving of the two directions for a
deployment to be wrong in. This is also the practical reading of the disclosure
argument in Section~\ref{sec:deployment}. A retention policy that permits one or
two records per old class, together with an operating history that has produced
three examples of each new condition, is enough to admit the site.

\section{Per-split grouping outcomes}
\label{app:seeds}

Table~\ref{tab:seed_diagnostics} reports the grouping stage for each of the
\nSeedsEtoE{} complete runs in Table~\ref{tab:controlled}(b). ``Groups
recovered'' is the number of conditions the clustering returned, and the
partition agreement with the reference conditions varies across splits.
``Unassigned'' counts flagged records that DBSCAN left in no group; these are
excluded from the new-class training set, which is why a run can recover the
correct number of groups and still train on fewer records than it flagged.

\begin{table}[ht]
\caption{Grouping outcomes for the complete runs. ARI is the adjusted Rand index
between the recovered partition and the reference conditions, computed on the
assigned records.}
\label{tab:seed_diagnostics}
\centering
\small
\begin{tabular}{lccc}
\toprule
Seed & Groups recovered & ARI & Unassigned \\
\midrule
7 & 3 & 0.764 & 12 \\
11 & 3 & 0.507 & 9 \\
42 & 3 & 0.734 & 11 \\
123 & 3 & 0.679 & 13 \\
314 & 3 & 0.667 & 9 \\
\bottomrule
\end{tabular}

\end{table}

\section{Reproducibility and compute}
\label{app:reproducibility}

Experiments ran with Python 3.11, TensorFlow 2.15.1 (Keras 2.15), NumPy 1.26.4,
pandas 1.5.3, scikit-learn 1.3.2, and Matplotlib 3.7.5, on two CPU cores with no
GPU. One complete chained run covering all configurations takes roughly ten
minutes, one controlled sweep over the seven subsets about five, and one seed of
the reference-method sweep about five, and one seed of the budget grid in
Appendix~\ref{app:budget} about six. We re-ran the base federated session after
every environment change and obtained identical end-of-session accuracy, with
later stages agreeing to within one test observation.

Randomness is controlled for NumPy and TensorFlow operations. Reported variation
is the sample standard deviation across independently seeded runs, used
descriptively; we draw no inferential conclusion from it.

The base classifier uses ReLU activations and a softmax output and is trained for
20 federated rounds with one local epoch per round, batch size four, and Adam at
$10^{-3}$. Each denoising autoencoder has hidden width 64, latent width 32,
Gaussian input noise of standard deviation 0.02, and $\ell_2$ coefficient
$10^{-4}$, trained for 30 federated rounds with two local epochs. Grouping uses
four principal components, and DBSCAN six minimum neighbors with a radius at the
60th percentile of empirical nearest-neighbor distances. Adaptation
runs at batch size four and learning rate $10^{-3}$.

\clearpage
\section{Notation}
\label{app:notation}

\begin{table}[ht]
\centering
\small
\begin{tabular}{ll}
\toprule
Symbol & Meaning \\
\midrule
$M$, $K$ & number of established sites; number of old classes \\
$R$ & number of new conditions recovered and added as outputs \\
$\mathcal{U}$ & unlabeled pool held by the joining site \\
$\phi_\theta(x)$ & early-layer activation, shared by all stages \\
$g_c$, $e_c(x)$ & reconstruction expert for class $c$; its squared error \\
$s_c$ & scale of expert $c$, its median error on its own class \\
$S(x)$ & screening score, $\min_c e_c(x)/s_c$ \\
$\tau$, $q$ & federated threshold; detection quantile \\
$p_{m,c}$ & prototype of class $c$ at established site $m$ \\
$b$ & local rehearsal records kept per old class \\
$\lambda_{\mathrm{p}}$ & weight on the prototype replay loss \\
$A_{\old}^{0}$, $A_{\old}$ & old-class accuracy before and after integration \\
$A_{\newc}$ & new-class accuracy after integration \\
$F_{\old}$ & forgetting, $A_{\old}^{0}-A_{\old}$ \\
\bottomrule
\end{tabular}
\end{table}

\end{document}